\documentclass[letterpaper, 10 pt, conference]{ieeeconf}  % Comment this line out if you need a4paper
\IEEEoverridecommandlockouts                              % This command is only needed if
\usepackage{graphicx} % for pdf, bitmapped graphics files
\usepackage{times} % assumes new font selection scheme installed
\usepackage{cite}
\usepackage{amsmath} % assumes amsmath package installed
\usepackage{amssymb}  % assumes amsmath package installed
\usepackage{amsfonts}       % blackboard math symbols
\usepackage{dsfont}
\usepackage{braket}
\usepackage{booktabs}
\usepackage{color}
\usepackage{colortbl}
\usepackage{url}
\usepackage{mathtools}
\usepackage{multirow}
\usepackage{footmisc}
\usepackage{mwe}
\usepackage[linesnumbered,ruled,vlined]{algorithm2e}
\SetKwInput{KwInput}{Input}                % Set the Input
\SetKwInput{KwOutput}{Output}

\SetCommentSty{mycommfont}

\usepackage{xcolor}

\usepackage[caption=false, font=footnotesize]{subfig}

\usepackage{lipsum}
\usepackage{widetext}

\newcommand{\traj}{\tau}
\newcommand{\ftraj}{\mathcal{F}}
\newcommand{\obs}{\boldsymbol{x}}
\newcommand{\obsattr}{\obs^{\mathrm{attr}}}
\newcommand{\trajdemo}{\traj^{\mathrm{demo}}}
\newcommand{\trajattr}{\traj^{\mathrm{attr}}}
\newcommand{\ftrajdemo}{\ftraj^{\mathrm{demo}}}
\newcommand{\ltask}{\mathcal{L}_{\mathrm{task}}}
\newcommand{\lcomp}{\mathcal{L}_{\mathrm{comp}}}

\newcommand{\vel}{\dot{\obs}}
\newcommand{\inertia}{\boldsymbol{\Lambda}}
\newcommand{\dampcoef}{\mathbf{D}}
\newcommand{\stiffness}{\mathbf{K}}
\newcommand{\stiffnesstraj}{\mathcal{K}}
\newcommand{\force}{\boldsymbol{F}}

\title{\LARGE \textbf{
  A Contact Model based on Denoising Diffusion to \\ Learn Variable Impedance Control for Contact-rich Manipulation
}}

\author{Masashi Okada$^{\dag,\star}$, Mayumi Komatsu$^{\ddag}$ and Tadahiro Taniguchi$^{\dag,*}$% <-this % stops a space
\thanks{$^{\dag}$ Masashi Okada and Tadahiro Taniguchi are with Digital \& AI Technology Center, Technology Division, Panasonic Holdings Corporation, Japan.
}%
\thanks{$^{\ddag}$ Mayumi Komatsu is with Robotics Promotion Office, Manufacturing Innovation Division, Panasonic Holdings Corporation, Japan.
}%
\thanks{$^{*}$ Tadahiro Taniguchi is also with Ritsumeikan University, College of Information Science and Engineering, Japan.
}%
\thanks{$^{\star}$ \texttt{okada.masashi001@jp.panasonic.com}
}
}

\begin{document}

\maketitle
\thispagestyle{empty}
\pagestyle{empty}

%%%%%%%%%%%%%%%%%%%%%%%%%%%%%%%%%%%%%%%%%%%%%%%%%%%%%%%%%%%%%%%%%%%%%%%%%%%%%%%%
\begin{abstract}
In this paper, a novel approach is proposed for learning robot control in contact-rich tasks such as wiping, by developing Diffusion Contact Model (DCM).
Previous methods of learning such tasks relied on impedance control with time-varying stiffness tuning by performing Bayesian optimization by trial-and-error with robots.
The proposed approach aims to reduce the cost of robot operation by predicting the robot contact trajectories from the variable stiffness inputs and using neural models.
However, contact dynamics are inherently highly nonlinear, and their simulation requires iterative computations such as convex optimization. 
Moreover, approximating such computations by using finite-layer neural models is difficult.
To overcome these limitations, the proposed DCM used the denoising diffusion models that could simulate the complex dynamics via iterative computations of multi-step denoising, thus improving the prediction accuracy.
Stiffness tuning experiments conducted in simulated and real environments showed that the DCM achieved comparable performance to a conventional robot-based optimization method
while reducing the number of robot trials.
\end{abstract}

%%%%%%%%%%%%%%%%%%%%%%%%%%%%%%%%%%%%%%%%%%%%%%%%%%%%%%%%%%%%%%%%%%%%%%%%%%%%%%%%
\section{Introduction} \label{sec:intro}
Learning from demonstration, particularly robot control through the playback of demonstrated trajectories, is a standard technology in the industry.
Such controls can be applied to contact-rich tasks using impedance control and careful tuning of its parameters (i.e., time-variant stiffness) to optimize the reproducibility of demonstration (i.e., \textit{task objective}) and safety against unexpected contacts (i.e., \textit{compliance objective})~\cite{pollayil2022choosing,okada2023learning}.
In tasks wherein robots have to acquire various skills, tuning must be completed in a short time, ideally in only single demonstration and a few tests.
For instance, robotic wiping requires skills to adapt various curved surfaces and trajectories.
\begin{figure}[t]
  \centering
  \includegraphics[width=0.48\textwidth]{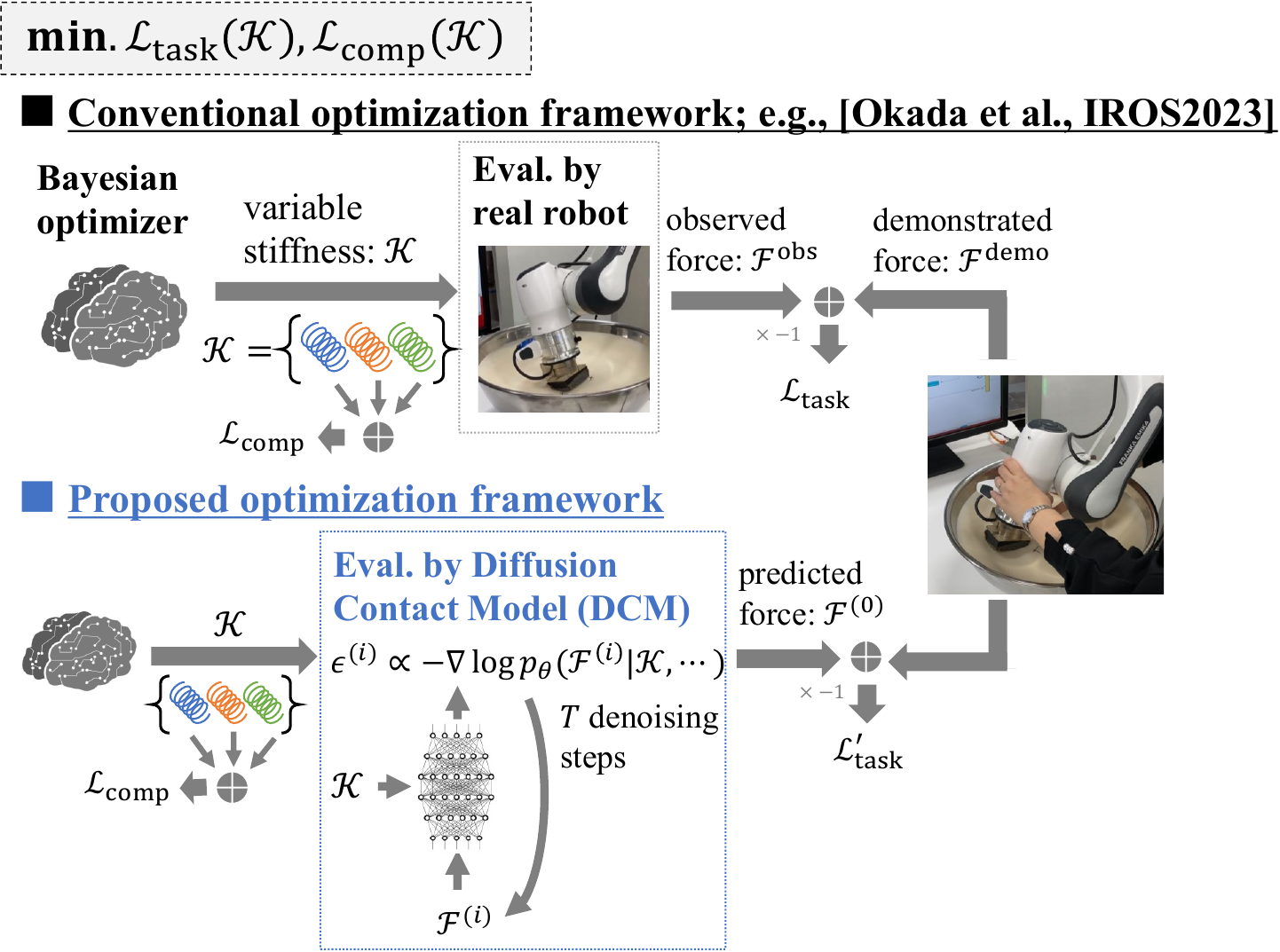}
  \caption{
    The concept of learning of variable stiffness $\stiffnesstraj$ formulated as a multi-objective optimization of the \textit{task objective} $\ltask$ and \textit{compliance objectives} $\lcomp$. 
    \textbf{(Top) Conventional framework (or robot-based optimization):} 
    the task objective is evaluated using a real robot~\cite{johannsmeier2019framework,wu2022prim,8566177,salehi2008impedance,fateh2011adaptive,azimi2015stable,okada2023learning}.
    \textbf{(Bottom) Proposed framework (or robot-free optimization):} the task objective is evaluated using the proposed diffusion contact model (DCM) without robot trials.
  }
  \label{fig:fig1}
\end{figure}

In previous literature on contact-rich impedance control, the stiffness was determined using black-box optimization such as Bayesian optimization for the robots to acquire new skills~\cite{johannsmeier2019framework,wu2022prim,8566177,salehi2008impedance,fateh2011adaptive,azimi2015stable,okada2023learning}.
Particularly, Ref.~\cite{okada2023learning} utilized multi-objective optimization to find optimal solutions that represent best tradeoff between the \textit{task and compliance objectives}.
However, multiple trials have to be performed with real robots (e.g., a few hours of robot operations), which prevent the robots from acquiring a variety of skills.

Robots cannot acquire new skills efficiently without the use of models that accurately simulate contact.
If a model that precisely simulates robot behavior with contact is developed,
model-based stiffness tuning can be performed without robot trials.
However, even in state-of-the-art robot simulators such as MuJoCo~\cite{todorov2012mujoco} and PyBullet~\cite{coumans2016pybullet}, simulated-to-real (sim2real) gap exists in contact dynamics~\cite{lidec2023contact}.
In addition, environment models must be constructed for simulating the geometry, friction, elastic properties of the robots and objects; however, constructing such models is difficult.

Contact simulation is difficult due to the high nonlinearity of contact dynamics.
Existing robot simulators \cite{todorov2012mujoco,coumans2016pybullet,macklin2019small,8255551,lee2018dart,tedrake2019drake,smith2005open} formulate contact simulation as a nonlinear complementarity problem (NCP)~\cite{acary2018solving} and solve it via iterative calculations such as convex optimization.
Approximating this nonlinearity with neural networks seems a promising approach, and it is also expected to bridge the sim2real gap by exploiting real data.
However, deep architectures with extensive parameters are required for the neural modeling of complex dynamics, which need a significant amounts of training data to avoid model overﬁtting. 

Herein, a diffusion contact model (DCM) is proposed to solve the aforementioned problems.
DCM employs denoising diffusion models~\cite{ho2020denoising},
a cornerstone in the recent advancements in \textit{generative AI},
for contact dynamics modeling.
Diffusion models were used because they have high nonlinearity due to multi-step denoising.
This iterative denoising can simulate deep networks but with fewer parameters, thus avoiding overfitting.
In addition, the denoising process acts as an optimization in a learned objective function~\cite{song2020score}.
In other words, multi-step denoising and iterative computation for contact simulation (e.g., convex optimization) are similar in terms of optimization, suggesting that DCM can learn the intrinsic process of contact simulation in a data-driven manner.

In this paper, we verify the effectiveness of DCM via simulation and actual experiments of learning robotic wiping as a target task.
This task requires a robot to reproduce the demonstrated contact forces between the end-effector and curved surfaces.
Figure~\ref{fig:fig1} summarizes the overall concept.
This paper follows the framework of a previous study~\cite{okada2023learning} to formulate variable impedance control learning as a multi-objective optimization of the task and compliance objectives.
% However, we can successfully reduce the number of trials on the real robot for the task objective evaluation by utilizing the proposed DCM.
The main contributions of this paper are as follows:
\begin{itemize}
  \item We introduce DCM for predicting the contact force trajectories of robots from variable stiffness input via iterative denoising based on diffusion models. 
  \item We demonstrate the superiority of DCM compared to a conventional neural model in terms of prediction accuracy of simulated and real wiping trajectories. %of the Franka Emika Panda robot arm.
  \item We also show that DCM can perform \textit{robot-free} optimization with comparable performance as that with \textit{robot-based} optimization via robotic wiping experiments.
\end{itemize}

The remainder of this paper is organized as follows.
Section~\ref{sec:related_work} summarizes related work.
Section~\ref{sec:preliminary} briefly reviews the preliminaries, i.e., impedance control,  variable impedance control learning as a multi-objective optimization, and denoising diffusion models.
Section~\ref{sec:method} presents DCM, and Sec.~\ref{sec:experiments} demonstrates its effectiveness. %via simulated evaluations.
Finally, Sec.~\ref{sec:conclusion} concludes the paper.

\section{Related Work} \label{sec:related_work}
\subsection{Contact Models in Robotics} \label{sec:contact_models_in_robotics}
Several models have been proposed for estimating the contact forces between objects, such as
the linear complementarity problem and the cone complementarity problem (CCP),
which are used by off-the-shelf simulators such as
MuJoCo~\cite{todorov2012mujoco}, Bullet~\cite{coumans2016pybullet}, PhysX~\cite{macklin2019small}, RaiSim~\cite{8255551},
DART~\cite{lee2018dart}, Drake~\cite{tedrake2019drake}, and ODE~\cite{smith2005open}.
Each of these methods approximates the NCP of contacts~\cite{acary2018solving}, but they commonly require iterative calculations.
In particular, MuJoCo reformulates the CCP as a convex optimization problem and solves it using  a quadratic programming solver.

\subsection{Robot Behavior Prediction using Neural Models} \label{sec:neural_models_in_robotics}
In the model-based reinforcement learning literature,
neural networks are commonly used to predict the future state of robots.
Recent studies have shown that the neural models can predict not only robot proprioception
\cite{nagabandi2018neural,chua2018deep,nagabandi2020deep,okada2020variational} 
but also high-dimensional vectors such as images
\cite{hafner2019dreamer,hafner2020mastering,
okada2020planet,wu2023daydreamer}.
However, few studies have focused on impedance control or contact force prediction~\cite{gao2020learning,anand2023model}.
A neural model was previously used~\cite{anand2023model} to perform variable impedance control with model predictive control; however external forces were regarded as actions therein (i.e., input to the neural model).
A recurrent neural model~\cite{gao2020learning} was also used to predict the contact force at the next timestep; however, it was not used for planning but for detecting anomalies by comparing the predicted and observed forces.

\subsection{Diffusion Models in Robotics}
Neural models in Sec.~\ref{sec:neural_models_in_robotics} predict future trajectories autoregressively, thereby accumulating step errors.
This issue can be addressed using diffusion models that simultaneously infer all timesteps~\cite{janner2022planning,ajay2022conditional,carvalho2023motion,mishra2023reorientdiff}.
Recently, diffusion models have been commonly used in reinforcement learning to represent policies and value functions~\cite{wang2022diffusion,hansen2023idql,chi2023diffusion}.

% RT-1,2 についても触れておく？~\cite{brohan2022rt,brohan2023rt}
% Optimization in Surrogate model~[引用]

\section{Preliminaries} \label{sec:preliminary}
\subsection{Impedance Control}
Impedance control imposes the robot's dynamics to follow the closed-impedance model:
\begin{align}
  & \inertia \Delta \ddot{\obs} + \dampcoef \Delta \vel + \stiffness \Delta \obs = \force \label{eqn:impedance_control}, \\
  & \Delta \obs = \obsattr - \obs,
\end{align}
where $\obs \in \mathbb{R}^{6}$ is the end-effector pose in task space,
$\obsattr \in \mathbb{R}^{6}$ is the attractor,
$\force \in \mathbb{R}^{6}$ is the external force/torque acting on the end-effector,
and the matrices
$\inertia \in \mathbb{R}^{6\times 6}$,
$\dampcoef \in \mathbb{R}^{6\times 6}$,
and $\stiffness \in \mathbb{R}^{6\times 6}$ are
the desired Cartesian inertia, damping, and stiffness, respectively.
In this context, stiffness $\stiffness$ that exhibits a spring-like behavior enables the robot to follow the desired trajectory of the attractor $\obsattr$ while maintaining flexibility in response to unexpected external forces $\force$.
To ensure system compliance, stiffness $\stiffness$ should be maintained low as possible.

\subsection{Formulation to Learn Variable Impedance Control}
To achieve safe and satisfactory task performance, variable impedance control is required using appropriately designed stiffness~\cite{7110619,7560657}.
For instance, let us reproduce a demonstrated trajectory $\trajdemo = \obs^{\mathrm{demo}}_{1:H}$, $\ftrajdemo = \force^{\mathrm{demo}}_{1:H}$ with variable stiffness $\stiffnesstraj = \stiffness_{1:H}$, where $H$ is the length of the trajectory.
The variable stiffness $\stiffnesstraj$ affects the safety and reproducibility of the demonstration, and a tradeoff exists the two objectives~\cite{pollayil2022choosing,okada2023learning}.
Thus, the stiffness learning problem can be formulated as the optimization of two objective functions:
\begin{align}
  \min_{\stiffnesstraj} \ltask(\stiffnesstraj), \lcomp(\stiffnesstraj), \label{eqn:multi_obj_opt}
\end{align}
where $\ltask$ is the \textit{task objective} and $\lcomp$ is the \textit{compliance objective}.
This study defines the objectives as follows:
\begin{align}
  \ltask(\stiffnesstraj) &\coloneqq \mathrm{RMSE}\left(\ftraj^{\mathrm{obs}}(\stiffnesstraj), \ftraj^{\mathrm{demo}}\right), \\
  \label{eqn:task_obj}
% \end{align}
% %
% \begin{align}
  \lcomp(\stiffnesstraj) &\coloneqq \sum_{t=1}^{H} |\stiffness_{t}|,
\end{align}
where $\ftraj^{\mathrm{obs}}(\stiffnesstraj)$ is the observed force on the end-effector.
As we focus on the reproduction of demonstrated forces, $\ltask$ is defined as the root-mean-squared error (RMSE) between the observed and demonstrated forces.
The damping $\dampcoef$ and attractor $\obsattr$ are determined using the following Eqs.~(\ref{eqn:critically_dumped}), (\ref{eqn:attractor}) \cite{okada2023learning}:
\begin{align}
  \dampcoef = 2 \stiffness^{\frac{1}{2}} \label{eqn:critically_dumped},
\end{align}
\begin{align}
  \obs^{\mathrm{attr}}_{t} = \obs^{\mathrm{demo}}_{t} + \stiffness^{-1}_{t} (\dampcoef \vel^{\mathrm{demo}}_{t} + \inertia \ddot{\obs}^{\mathrm{demo}}_{t} - \force^{\mathrm{demo}}_{t}), \label{eqn:attractor}
\end{align}
and are not included in the search space of the optimization problem.

The optimization of the variable stiffness $\stiffnesstraj$ for all timesteps is difficult due to the high dimensionality of the search space.
The number of dimensions was previously reduced~\cite{okada2023learning} by segmenting a trajectory into several phases and assigning constant responsible
stiffness values to each phase.
Herein, this segmentation method was used for learning stiffness.
The details of the method are beyond the scope of this study and are therefore omitted.

\begin{figure}[t]
  \centering
  \includegraphics[width=0.35\textwidth]{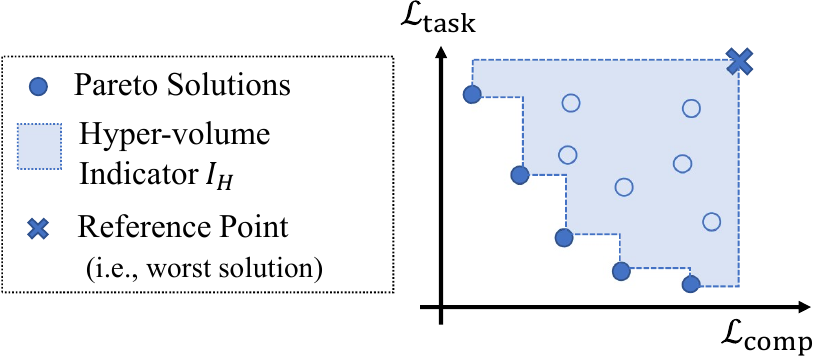}
  \vspace*{-3mm}
  \caption{
    Illustration of Pareto solutions.
  }
  \label{fig:pareto_example}
\end{figure}

The multi-objective optimization problem in Eq.~(\ref{eqn:multi_obj_opt}) was solved to obtain several optimal \textit{Pareto solutions}, indicating the best tradeoff between the two objectives.
To quantitatively compare the solutions obtained using different methods,
the hypervolume indicator $I_{H}$ was used in the experiments in Sec.~\ref{sec:experiments}, which is defined by the area composed of Pareto solutions and a reference point.
Figure~\ref{fig:pareto_example} illustrates Pareto solutions and the hypervolume indicator.
% Fig.~\ref{fig:pareto_example} illustrates Pareto solutions and the hypervolume indicator are  in 
%

\subsection{Diffusion Models} \label{sec:diffusion_models}

% [TODO C\&P; to be revised]
For a dataset $\mathcal{D}$ wherein samples $\traj$ are drawn from an unknown data distribution $q(\traj)$%
\footnote{
  In this section, $\traj$ represents an arbitrary tensor.
}%
, diffusion models are used to approximate the distribution using a parameterized generative model $p_{\theta}(\traj)$.
The procedure is completed in two steps: forward diffusion and backward denoising.
Forward diffusion generate latents $\traj^{(1:T)}$ by stepwise injection of Gaussian noise into $\traj=\traj^{(0)}$ based on the following transitions:
$q(\traj^{(i)}|\traj^{(i-1)}) = \mathcal{N}(\traj^{(i)}; \sqrt{1-\beta_{i}}\traj^{(i-1)}, \beta_{i}I)$
with variance schedule $\beta_{1:T}$, yielding the distribution:
\begin{align}
  q(\traj^{(i)}|\traj^{(0)}) = \mathcal{N}(\traj^{(i)}; \sqrt{\bar{\alpha}_{i}}\traj, (1-\bar{\alpha}_{i})I), \label{eqn:noise_injection}
\end{align}
where
$\alpha_{i} \coloneqq 1 - \beta_{i}$, 
and
$\bar{\alpha}_{i} \coloneqq \prod_{i=1}^{T} \alpha_{i}$.
Backward denoising removes the injected Gaussian noise starting from $\traj^{(T)} \sim \mathcal{N}(0, I)$ based on the following denoising transition:
\begin{align}
  p_{\theta}(\traj^{(i-1)}|\traj^{(i)}) = \mathcal{N}(\traj^{(i-1)}; \mu_{\theta}(\traj^{(i)}, i), \beta_{i}I), \label{eqn:backward_denoising}
\end{align}
where
\begin{align}
  \mu_{\theta}(\traj^{(i)}, i) = \frac{1}{\sqrt{\alpha_{i}}}\left(
    \traj^{(i-1)} - \frac{\beta_{i}}{\sqrt{1-\bar{\alpha}_{i}}}\epsilon_{\theta}(\traj^{(i)}, i)
  \right).
\end{align}
The  parameterized  model $\epsilon_{\theta}(\traj^{(i)}, i)$ is known as \textit{the score-function}, using which the injected Gaussian noise by Eq.~(\ref{eqn:noise_injection}) can be predicted; it is trained by \textit{the implicit score-matching objective}~\cite{song2020score}:
\begin{align}
  & \mathcal{L}(\theta) =
  \mathbb{E}_{i, \traj^{(i)}\sim q(\traj^{(i)}|\traj^{(0)}), \traj^{(0)}\sim \mathcal{D}} \left[
\|\epsilon - \epsilon_{\theta}(\traj^{(i)}, i) \|^{2}
\right] \nonumber \\
  &= \mathbb{E}_{i, \epsilon\sim\mathcal{N}(0,I), \traj^{(0)}} \left[
    \|\epsilon - \epsilon_{\theta}(\sqrt{\bar{\alpha}_{i}}\traj^{(0)} + \sqrt{1-\bar{\alpha}_{i}}\epsilon, i) \|^{2} 
  \right] \label{eqn:score_matching},
\end{align}
Specifically, this score function contains the gradients of the learned probability distribution as follows:
\begin{align}
  \epsilon_{\theta}(\traj, i) \propto - \nabla_{\traj} \log p_{\theta}(\traj).\
  \label{eqn:gradients}
\end{align}

\section{Method} \label{sec:method}
\subsection{Diffusion Contact Model}
In this section, we introduce Diffusion Contact Model (DCM) whose concept is illustrated in Fig.~\ref{fig:denoising}.
% that predicts the contact force trajectory $\ftraj$ between robot's end-effector and surfaces from control parameter inputs
% based on diffusion models introduced in Sec.~\ref{sec:diffusion_models}.
% Figure~\ref{fig:denoising} illustrates DCM; e.g., the block-diagram, examples of input/output of the model.

\subsubsection{Formulation}
\begin{figure*}[t]
  \centering
  \includegraphics[width=0.95\textwidth]{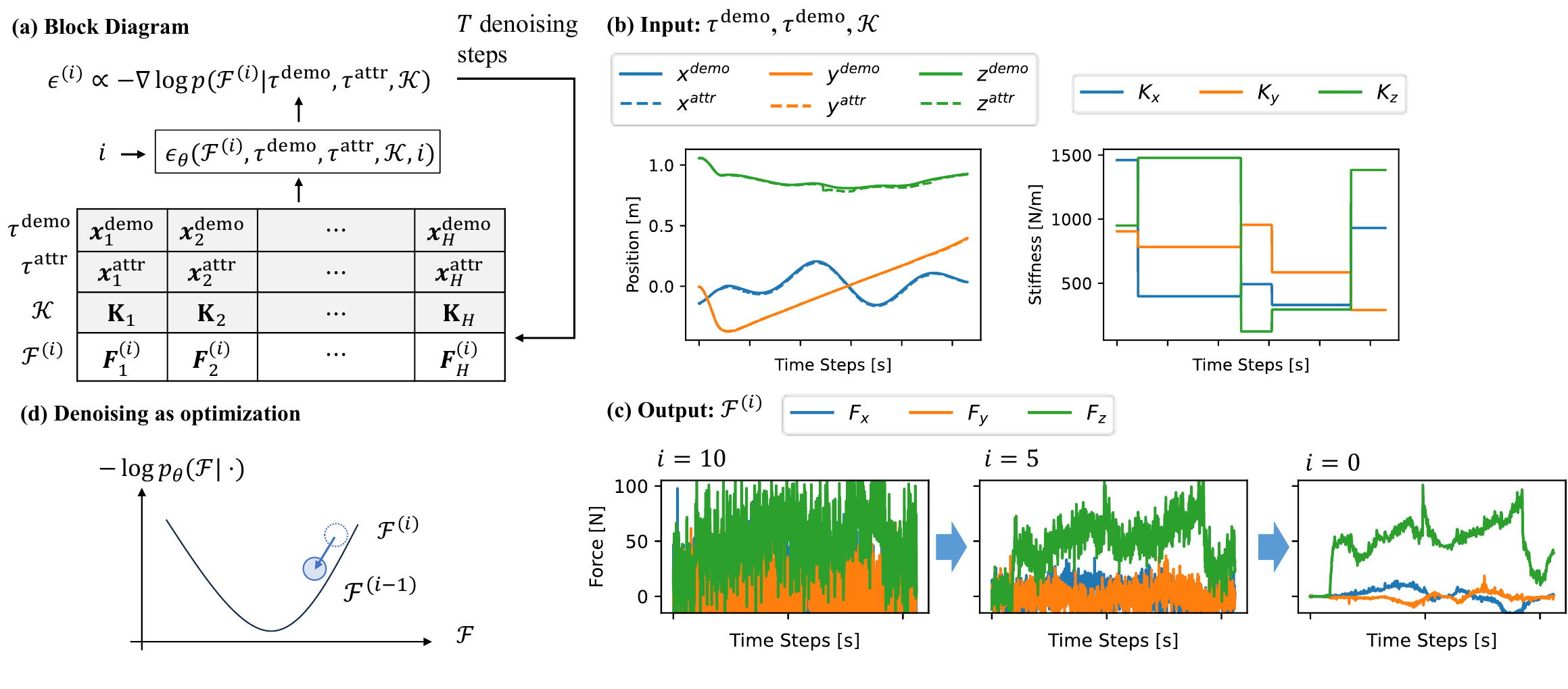}
  \vspace*{-2mm}
  \caption{
    Conceptual illustration of the diffusion contact model (DCM).
    \textbf{(a) Block-diagram:} DCM iteratively estimates the contact force trajectory $\ftraj$ by $T$ steps of denoising, from the input containing the demonstrated trajectory $\trajdemo$, reference trajectory $\trajattr$, and variable stiffness $\stiffnesstraj$.
    \textbf{(b/c) Examples of input/output}, in which the orientation-related sequences are omitted for visibility.
    \textbf{(d) Illustration of iterative denoising process.} 
    % Score function implementation by RetNet~\cite{Sun2023-pp}.
    % \texttt{embed(i)} denotes the embedding of the denoising step $i$ by Gaussian Fourier Projection~\cite{song2020score}.
    % \texttt{Tokenizer} is realized by a fully-connected layer that projects a concatenated input vectors to a token $\mathbf{e}_{t}$.
  }
  \label{fig:denoising}
\end{figure*}

DCM defines the prediction model as a conditional distribution of $p_{\theta}(\ftraj | \trajdemo, \trajattr, \stiffnesstraj)$, where $\trajdemo$ is the demonstrated trajectory, $\trajattr = \obsattr_{1:H}$ is the reference trajectory of the end-effector to be tracked, and $\stiffnesstraj$ is the variable stiffness.
$\stiffnesstraj$ is suggested by a Bayesian optimization algorithm as a solution candidate,
and $\trajattr$ is calculated from Eq.~(\ref{eqn:attractor}).
The demonstrated trajectory $\trajdemo$ implicitly provides the geometry of the contact surfaces, which is essential for predicting contact forces.
The prediction model comprises per-step denoising process $p_{\theta}\left({\ftraj^{(i-1)} | \ftraj^{(i)}, (\trajdemo, \trajattr, \stiffnesstraj)}\right)$ and score function $\epsilon_{\theta}\left(\ftraj^{(i)}, (\trajdemo, \trajattr, \stiffnesstraj), i\right)$.
Differing from Eq.~(\ref{eqn:backward_denoising}), $p_\theta(\ftraj^{(i-1)}|\cdot)$ and $\epsilon_{\theta}(\cdot)$ are conditioned on $(\trajdemo, \trajattr, \stiffnesstraj)$.
As shown in Fig.~\ref{fig:denoising}(a), this score function is implemented as a neural network that considers the condition variables $(\trajdemo$, $\trajattr$, $\stiffnesstraj)$ and $\ftraj^{(i-1)}$ as inputs.
Figs.~\ref{fig:denoising}(b,c) illustrates the input and output of DCM, respectively.

% Figure~\ref{fig:denoising} illlustrates the prediction procedure.
% To predict the prediction of contact forces, the model $p_{\theta}(\ftraj)$ needs to be conditioned on the control parameters $\stiffnesstraj$ and $\trajattr$.
% In addition, the model is conditioned on the demonstrated trajectory $\trajdemo$
% which implicitly includes the geometry of the contact surface.
% As a result, the prediction model is formulated as $p_{\theta}(\ftraj | \trajdemo, \trajattr, \stiffnesstraj)$, which comprises the per-step denoising process $p_{\theta}({\ftraj^{(i-1)} | \ftraj^{(i)}, \trajdemo, \trajattr, \stiffnesstraj})$ and score function $\epsilon_{\theta}(\ftraj^{(i)}, \trajdemo, \trajattr, \stiffnesstraj, i)$. 
% Figure~\ref{fig:denoising} illlustrates the prediction procedure.
%

\subsubsection{Dataset and Training}
We suppose that conventional Bayesian optimization (the top of Fig.~\ref{fig:fig1}) is applied and operated for a certain period using real robots, thus yielding robot operation logs that can be used as a dataset $\mathcal{D} \ni \{(\trajdemo, \trajattr, \stiffnesstraj), \ftraj\}$ to train the score function $\epsilon_{\theta}(\cdot)$ by optimizing Eq.~(\ref{eqn:score_matching}).

\subsubsection{Design of the Score Function}
\begin{figure}[t]
  \centering
  \includegraphics[width=0.45\textwidth]{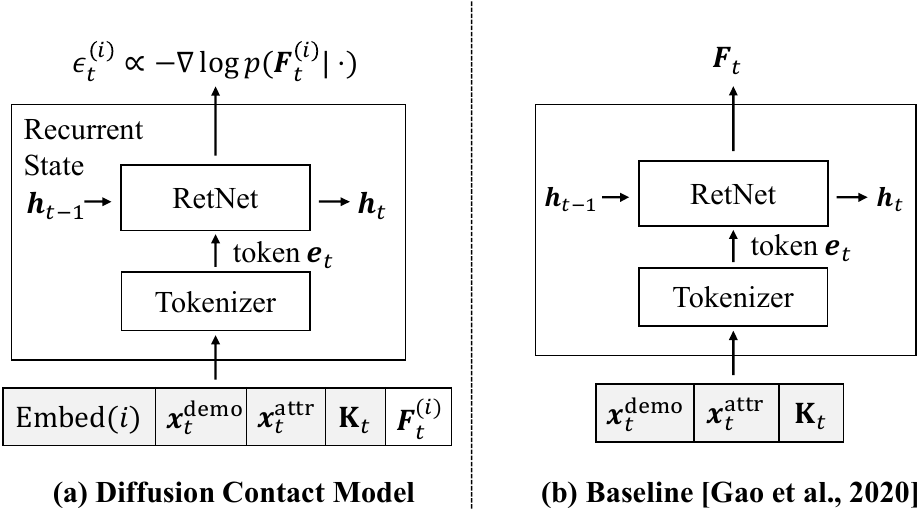}
  \vspace*{-3mm}
  \caption{
    \textbf{(a) Score function implementation} by a state-of-the-art recurrent neural model, RetNet~\cite{Sun2023-pp}.
    \texttt{Embed(i)} denotes the embedding of the denoising step $i$ by the Gaussian Fourier projection~\cite{song2020score}.
    \texttt{Tokenizer} is realized by a fully-connected layer, which projects a concatenated input vector to the token $\mathbf{e}_{t}$.
    The number of dimensions of $\mathbf{e}_{t}$ was set as $128$.
    \textbf{(b) Baseline model} used in Sec.~\ref{sec:experiments} which predicts forces in a single forward process (i.e., $T=1$).
    Bidirectional Gated Recurrent Units (Bi-GRU) were used previously~\cite{gao2020learning}; however RetNet was used here instread.
  }
  \label{fig:retnet}
\end{figure}

A recurrent neural network (RNN) is used herein to process the time-series input of the score function $\epsilon_{\theta}(\cdot)$ (Fig.~\ref{fig:retnet}(a)), similar to a previous study on the contact force prediction~\cite{gao2020learning} (Fig.~\ref{fig:retnet}(b)).
However, we introduced Retentive Network (RetNet)~\cite{Sun2023-pp}, a state-of-the-art RNN architecture with performance comparable with that of Transformer~\cite{vaswani2017attention} while considerably reducing the computational costs.
During force prediction, we empirically determined that RetNet exhibited better performance than U-Net~\cite{janner2022planning} and Transformer~\cite{chi2023diffusion}, which are commonly used to implement score functions with time-series input.
Regarding the noise schedule defined by $\beta_{1:T}$, we use the linear schedule ~\cite{ho2020denoising}.
The results of preliminary experiment on the design decision are detailed in Appendix.~\ref{sec:ablation_study}.

%
% \begin{align}
%   p_{\theta}(\ftraj | \trajdemo, \trajattr, \stiffnesstraj) = 
%   \prod_{i=0}^{T}
%   p_{\theta}({\ftraj^{(i-1)} | \ftraj^{(i)}, \trajdemo, \trajattr, \stiffnesstraj})
% \end{align}

\subsubsection{Learning to Simulate Contact via Iterative Denoising} \label{sec:motivation}
As can be seen in Eq.~(\ref{eqn:gradients}) and Fig.~\ref{fig:denoising}(d),
the prediction of $\ftraj$ using DCM can be interpreted as the following optimization:
\begin{align}
  \min_{\ftraj}\ - \log p_{\theta}( \ftraj | \trajdemo, \trajattr, \stiffnesstraj) \label{eqn:dcm_opt},
\end{align}
where the objective is automatically and implicitly acquired by learning to predict contact forces.
A formuation to simulate contact dynamics can be described as the convex optimization as follows~\cite{todorov2012mujoco}:
\begin{align}
  \min_{\ddot{\obs}}\  & (\inertia \ddot{\obs} - \force_{\mathrm{actuated}})^\top \inertia^{-1} (\inertia \ddot{\obs} - \force_{\mathrm{actuated}}) \nonumber \\
                      & + (\mathbf{J} \ddot{\obs} - \ddot{\obs}^{*}) \mathbf{R}^{-1} (\mathbf{J} \ddot{\obs} - \ddot{\obs}^{*}) \label{eqn:contact_opt},
\end{align}
where $\ddot{\obs}$ is a target variable (i.e., acceleration constrained by contact surfaces), $\ddot{\obs}^{*}$ is reference acceleration, $\force_{\mathrm{actuated}}$ is the forces actuated to robots, $\mathbf{J}$ is the Jacobian, and $\mathbf{R}$ is the matrix to regularize the constraint: $\mathbf{J} \ddot{\obs} = \ddot{\obs}^{*}$.
% このように DCM と接触力のシミュレーションは、最適化という観点で共通する
As can be seen in Eqs.~(\ref{eqn:dcm_opt}) and (\ref{eqn:contact_opt}), DCM and contact simulation share the same perspective of optimization, suggesting that DCM can acquire the objective $- \log p_{\theta}(\ftraj|\cdot)$ which simulates the objective similar to Eq.~(\ref{eqn:contact_opt}) in a data-driven manner.
By incorporating such an \textit{inductive bias}~\cite{baxter2000model} for contact simulation, we expect that DCM can suppress overfitting and achieve higher prediction accuracy for unseen inputs.
%
% iterative calculations, by which we incorporated \textit{inductive bias}~\cite{baxter2000model} required for the contact simulation.
% , suggesting that DCM could learn the intrinsic process of contact simulation .

% The rationale behind employing diffusion models for force prediction lies in their ability to simulate the highly nonlinear dynamics of contact outlined in Sec.~\ref{sec:contact_models_in_robotics}.
% Modeling such the complex dynamics with neural networks necessitates the use of deep architecture; however, deep networks with extensive parameters need a significant amount of training data to avoid overﬁtting. In contrast, the diffusion models can simulate deep networks but with fewer parameters through their iterative denoising steps.
% Furthermore, as seen in Eq.~(10) and Fig.~\ref{fig:denoising}, the denoising process acts as an optimization process, resembling the convex optimization in current simulators,
% suggesting that DCM could learn the intrinsic process of contact simulation in a data-driven manner.
% This suggests that it is feasible to achieve a process that imitates the optimization conducted by the simulators.

\subsection{Multi-objective Bayesian Optimization by DCM}
For a demonstrated trajectory, we tune the variable stiffness $\stiffnesstraj$ 
to acquire the demonstrated skill by solving Eq.~(\ref{eqn:multi_obj_opt}),
As shown in Fig.~\ref{fig:fig1} (bottom), the task objective $\ltask$ is estimated using DCM without conducting trials on real robots.
To solve the multi-objective optimization in Eq.~(\ref{eqn:multi_obj_opt}), we introduce a state-of-the-art Bayesian optimization method proposed in a previous study~\cite{ozaki2020multiobjective}.

\section{Experiments} \label{sec:experiments}
We evaluated the effectiveness of the proposed method
via simulations and real robot experiments. 
In the subsequent section, the task and dataset settings are specified in
Sec.~\ref{sec:task_and_dataset}.
Then, the prediction accuracy determined via simulations and real experiments are discussed in 
Sec.~\ref{sec:pred_acc_sim} and \ref{sec:pred_acc_real}, respectively.
Finally, the effectiveness of DCM in stiffness learning experiments is demonstrated in Sec.~\ref{sec:robot_free_opt}.
\subsection{Task and Dataset Preparation} \label{sec:task_and_dataset}
\subsubsection{Simulated Environment} \label{sec:sim_dataset}
\begin{figure}[t]
  \centering
  \includegraphics[width=0.3\textwidth]{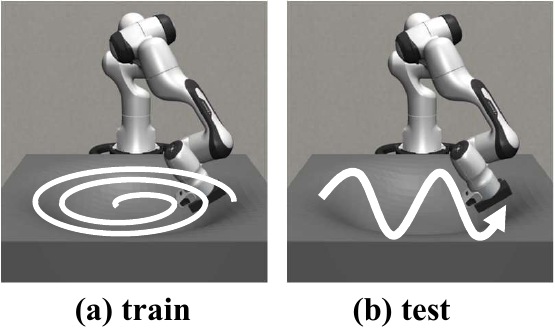}
  \vspace*{-2.5mm}
  \caption{
    Simulated wiping task environment and demonstrated trajectories.
    Demonstrations were performed by manually designed open-loop controllers.
    \textbf{(a) Trajectories for training:}
    we demonstrated 80 skills of spiral motion with different bowl sizes and directions,
    and then performed Bayesian optimization with $N=20$ trials to reproduce each demonstration, which yielded 1,600 trajectories.
    % Trials with anomalies (e.g., slips) occurred were eliminated.
    \textbf{(b) Trajectories for test:}
    we demonstrated a single skill of wavelike motion,
    and performed Bayesian optimization with $N=30$ trials.
  }
  \label{fig:sim_demos}
\end{figure}

We built a simulated environment for a wiping curved bowl task
using MuJoCo~\cite{todorov2012mujoco} and RoboSuite~\cite{robosuite2020}
(Fig.~\ref{fig:sim_demos}).
In the simulated environment, we collected 1,600 trajectories of spiral motions to train DCM and 30 trajectories of wavelike motions for accuracy evaluation and stiffness learning experiments.
The details of the dataset are summarized in the caption of Fig.~\ref{fig:sim_demos}.

\subsubsection{Real Environment} \label{sec:real_dataset}
\begin{figure}[t]
  \centering
  \includegraphics[width=0.3125\textwidth]{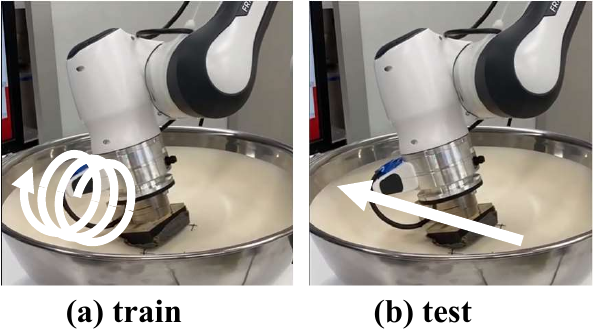}
  \vspace*{-2.5mm}
  \caption{
    Real wiping task environment and demonstrated trajectories.
    Demonstrations were performed by direct teaching.
    \textbf{(a) Trajectories for training:}
    we demonstrated four skills of spiral motion with different directions,
    and performed Bayesian optimization with $N=20$ trials, yielded 80 trajectories.
    \textbf{(b) Trajectories for test:}
    we demonstrated a single skill of line motion,
    and performed Bayesian optimization with $N=20$ trials.
    }
  \label{fig:real_demos}
\end{figure}

We also prepared a real environment of the wiping task using Franka Emika Panda, as shown in Fig.~\ref{fig:real_demos} for which 80 training trajectories and 20 test trajectories were collected.

\subsection{Prediction Accuracy on Simulation Dataset} \label{sec:pred_acc_sim}
We conducted this experiment to answer the following question:
\begin{itemize}
  \item Does DCM have higher prediction accuracy than conventional neural networks by incorporating iterative calculations?
\end{itemize}
To this end, training and evaluation were performed with varying diffusion steps $T$. Here, we regarded $T=1$ as the baseline method in Ref.~\cite{gao2020learning} and Fig.~\ref{fig:retnet}(b), which predicts contact forces in a single forward process of RNN.
Two metrics $\mathcal{L}_{\mathrm{MSE}}$ and $R$ were used to evaluate the prediction accuracy.
$\mathcal{L}_{\mathrm{MSE}}$ is the mean-squared error between the predicted and true forces for the test dataset%
\footnote{
We normalized all input and output to DCM to be in the range of $[-1, 1]$,
and computed $\mathcal{L}_{\mathrm{MSE}}$ using normalized values.
}.
$R$ is the correlation coefficient between the predicted and true values of the task objective $\ltask$ in Eq.~(\ref{eqn:task_obj}), which particularly affects the sebsequent Bayesian optimization.

\begin{figure}[t]
  \centering
  \includegraphics[width=0.425\textwidth]{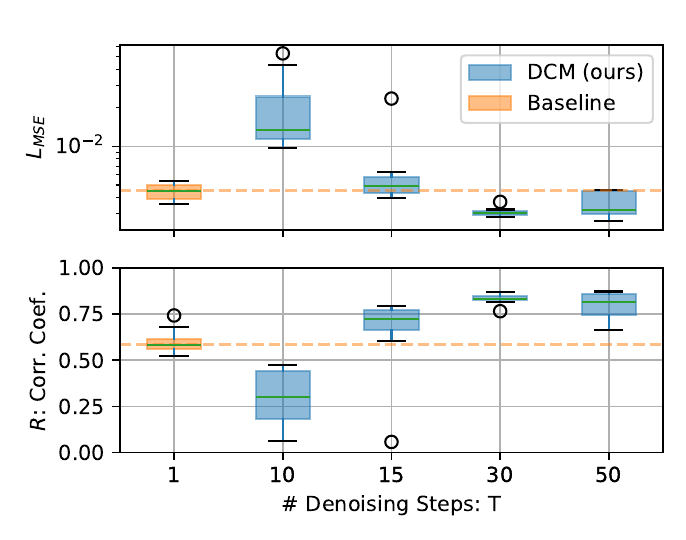}
  \vspace*{-5mm}
  \caption{
    Prediction accuracies for the simulated dataset with varying diffusion steps $T$,
    where $T=1$ represents the baseline method~\cite{gao2020learning}.
    Training was conducted eight times for each setting with different random seeds.
  } 
  \label{fig:pred_err}
\end{figure}

\begin{figure}[t]
  \centering
  \includegraphics[width=0.425\textwidth]{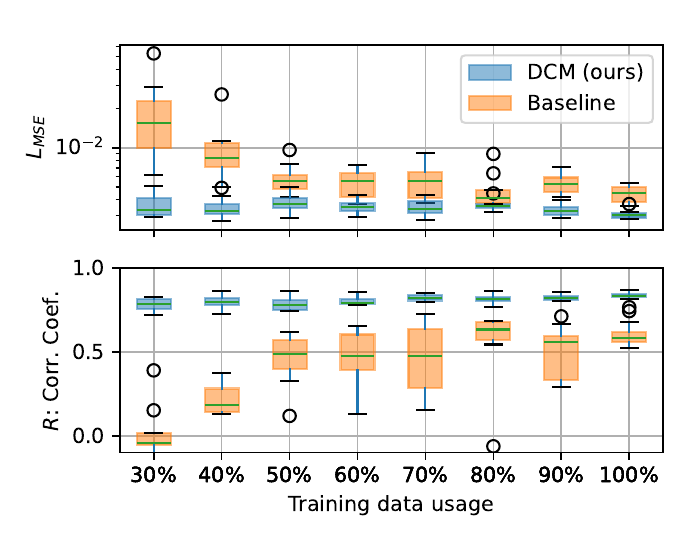}
  \vspace*{-5mm}
  \caption{
    Prediction accuracies for the simulated dataset with varying usage rate of training dataset, reporting the statistics from eight different seeds for each setting.
    For DCM, the diffusion steps $T$ was set as $30$.
    % DCM ($T=30$) achieved better prediction accuracy than the baseline (${p}\textrm{-val}=9.93\times10^{-6}$ for $\mathcal{L}_{\mathrm{MSE}}$).
  }
  \label{fig:ablation_dataset_size}
\end{figure}

\begin{figure}[t]
  \centering
  \includegraphics[width=0.45\textwidth]{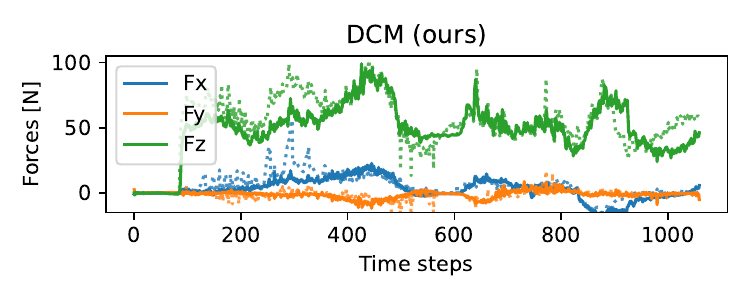}\vspace*{-8mm}
  \includegraphics[width=0.45\textwidth]{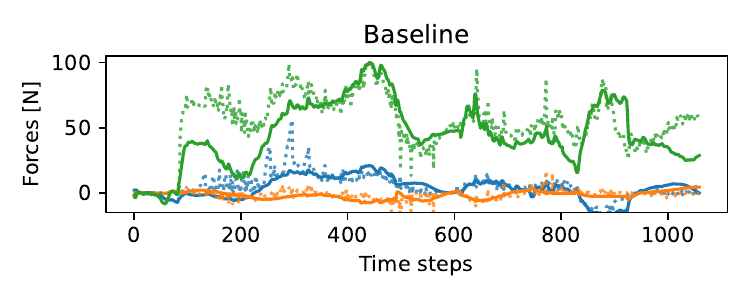}
  \vspace*{-5mm}
  \caption{
    Examples of the predicted contact force trajectories from \textit{simulated} test data input.
    Solid and dashed lines represent the predicted and true forces, respectively.
    $F_{x}$ and $F_{y}$ represent the tangential forces (i.e., friction), and $F_{z}$ represents the normal forces. 
    % The torques applied to the end-effector were also predicted but omitted here for visibility.
  }
  \label{fig:comp_pred_results}
\end{figure}

Figure \ref{fig:pred_err} shows the evaluation results of training using the simulated dataset, which demonstrates that DCM outperforms the baseline method (i.e., $T=1$) on two metrics at a certain step (i.e., $T=30$).
Although the performance improves as $T$ increases, the performance seems to saturate at $T=30$.
Figure \ref{fig:ablation_dataset_size} shows the effectiveness of DCM from another perspective, in which the prediction accuracy was evaluated using varying usage rates of the training dataset.
Although the baseline method and DCM have similar neural architectures, the prediction accuracy of the baseline method degrades rapidly as the usage rate decreases, whereas the performance of DCM remains intact.
As discussed in Sec.~\ref{sec:motivation}, this is probably because DCM with iterative calculations acquired the intrinsic procedure of contact dynamics simulation by learning to predict contact 
forces, thus suppressing overfitting.
Figure \ref{fig:comp_pred_results} shows examples of force trajectories predicted by DCM and baseline 


\subsection{Prediction Accuracy on Real Dataset} \label{sec:pred_acc_real}
This experiment was conducted to answer the following question:
\begin{itemize}
  \item Can DCM predict contact forces in real environments?
\end{itemize}
An issue with real environments is that sufficient training data are rarely available.
Therefore, the models pretrained with abundant simulated data were fine-tuned herein.

\begin{figure}[t]
  \centering
  \includegraphics[width=0.425\textwidth]{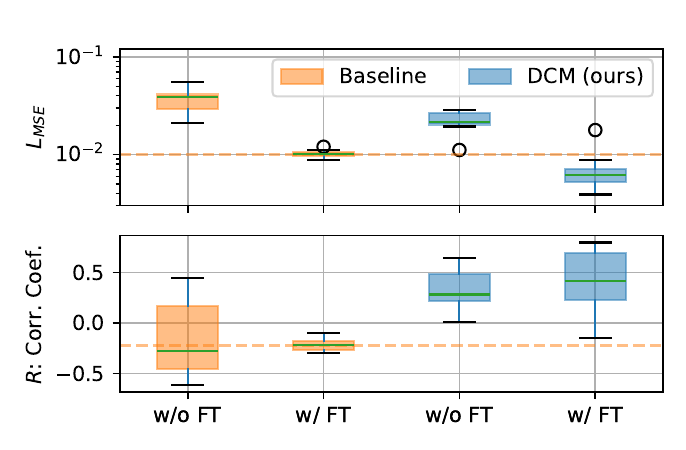} 
  \vspace*{-4mm}  
  \caption{
    Prediction accuracies for real dataset.
    \texttt{FT} denotes \textit{fine-tuning}.
    Training was conducted eight times for each setting with different random seeds.
    $T$ was set as 30 for DCM.
    % In the finetuning setting, DCM outperforms the baseline (${p}\textrm{-val}=3.24\times10^{-8}$ for $\mathcal{L}_{\mathrm{MSE}}$).
    }
  \label{fig:pred_err_real}
\end{figure}

\begin{figure}[t]
  \centering
  \includegraphics[width=0.45\textwidth]{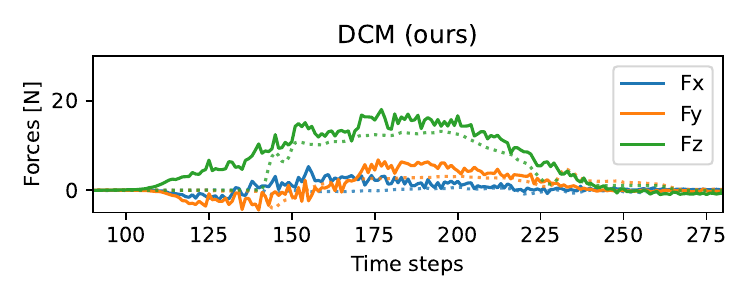}\vspace*{-8mm}
  \includegraphics[width=0.45\textwidth]{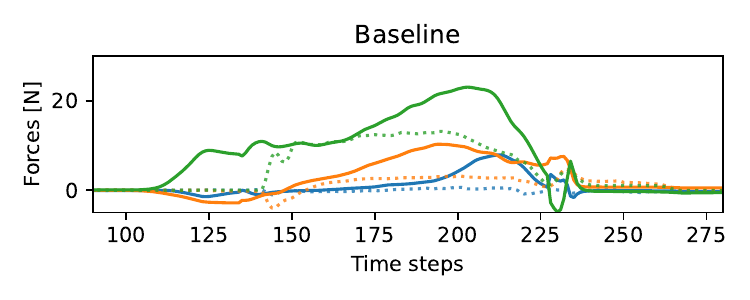}
  \vspace*{-5mm}
  \caption{
    Examples of predicted contact force trajectories from \textit{real} test data input.
    Solid and dashed lines represent the predicted and true forces, respectively.
  }
  \label{fig:comp_pred_results_real}
\end{figure}

DCM and baseline models were trained and tested using the dataset described in Sec.~\ref{sec:real_dataset} with and without fine-tuning, respectively.
Pretrained models were obtained from the previous section. %(the usage rate is 100\%.).
Fig.~\ref{fig:pred_err_real} summarizes the evaluation results, incicating that fine-tuned DCM exhibited the best performance.
Although the baseline model did not show any correlation ($R \simeq 0$) between the true and predicted values of $\ltask$, DCM showed a specific positive correlation ($R \simeq 0.5$), suggesting that the learned model can be used for Bayesian optimization.
Figure ~\ref{fig:comp_pred_results_real} shows the predicted force trajectories by DCM and baseline.

\subsection{Stiffness Learning using DCM} \label{sec:robot_free_opt}
We conducted stiffness learning experiments to verify the following question:
\begin{itemize}
  \item Can we conduct stiffness learning with DCM while reducing the number of robot trials?
\end{itemize}
%
% ここではシミュレーション、実機での test demonstration を再現するための stiffness を学習する。
The objective of stiffness learning was to reproduce the test demonstrations in the simulated and real environments.
Multi-objective Bayesian optimization (Eq.~(\ref{eqn:multi_obj_opt})) was performed, but the task objective $\ltask$ was estimated using DCM and the baseline model.
The number of virtual trials $N$ for the robot-free optimization was set as 1,000.
The Pareto solutions obtained via robot-free optimization were validated by conducting robot trials. Then the \textit{genuine} hypervolume indicators $I_{H}$ were compared.
The validity of the \textit{robot-free} optimization results was verified by comparing those with the \textit{robot-based} results (i.e., optimization with $N$ robot trials).

\begin{figure}[t]
  \centering
  \includegraphics[width=0.425\textwidth]{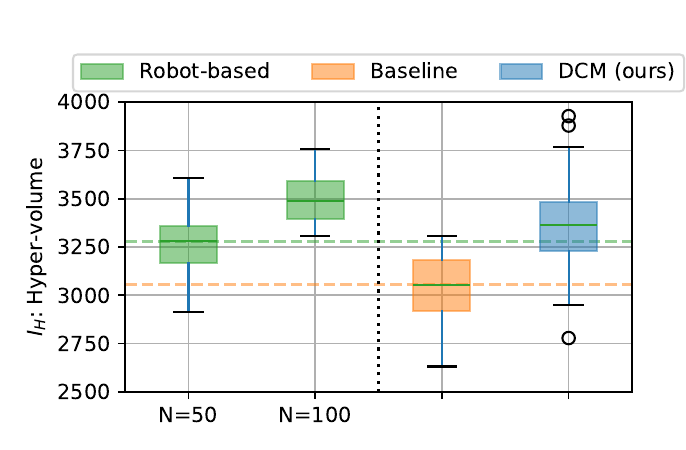}
  \vspace*{-4mm}
  \caption{
    Hypervolume indicators of the optimization results, where $N$ is the number of simulated robot trials.
    Experiments were conducted 64 times for each setting using different random seeds.
  }
  \label{fig:hypervolume}
\end{figure}

\begin{figure}[t]
  \centering
  \includegraphics[width=0.425\textwidth]{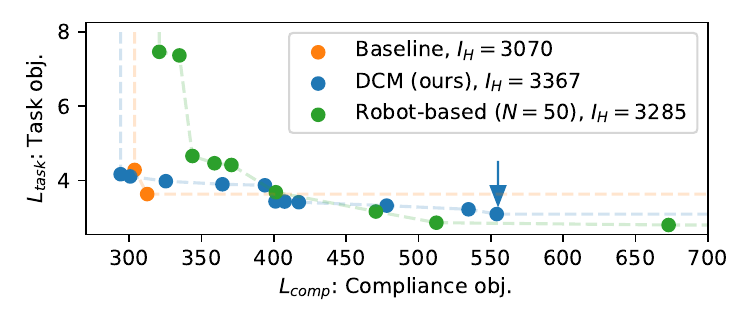}\vspace*{-3mm}
  \includegraphics[width=0.425\textwidth]{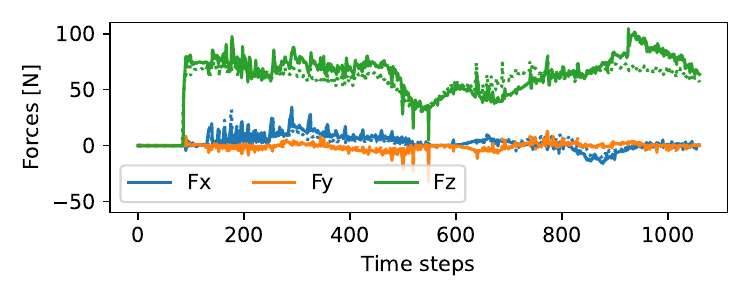}
  \vspace*{-5mm}
  \caption{
    Simulated stiffness learning results.
    \textbf{(Top) Pareto solutions} of robot-free (using baseline and DCM) and robot-based optimization.
    \textbf{(Bottom) Contact forces} realized by a solution found by DCM (indicated by an arrow in the top part). Solid and dashed lines represent the realized and demonstrated forces, respectively.
  }
  \label{fig:pareto_sim}
\end{figure}

Figure \ref{fig:hypervolume} compares the hypervolume indicators $I_{H}$ in the simulation setting, in which DCM outperformed the baseline method and was comparable to the robot-based optimization with $N=50$ trials, demonstrating the validity of robot-free optimization performed using DCM.
Figure.~\ref{fig:pareto_sim} shows the Pareto solutions and realized contact forces.

\begin{figure}[t]
  \centering
  \includegraphics[width=0.425\textwidth]{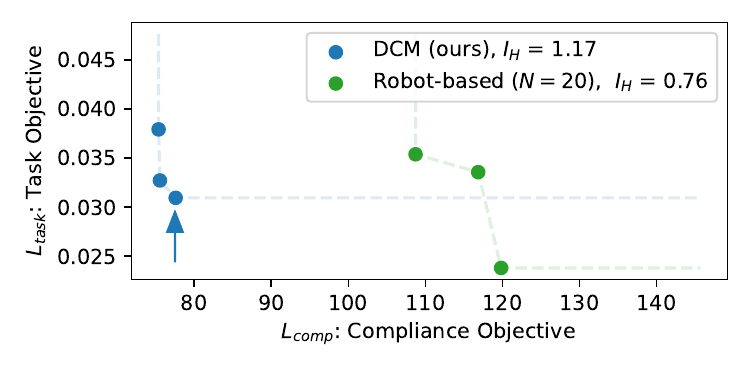}\vspace*{-3mm}
  \includegraphics[width=0.425\textwidth]{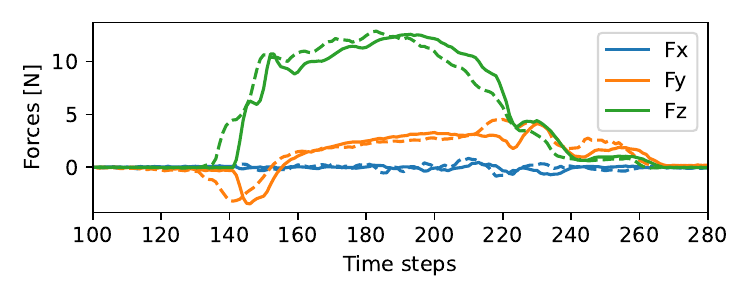}
  \vspace*{-5mm}
  \caption{
    Real robot stiffness learning results.
    \textbf{(Top) Pareto solutions} of robot-free (DCM) and robot-based optimization.
    \textbf{(Bottom) Contact forces} realized by a solution found by DCM (indicated by an arrow in the top part). 
  }
  \label{fig:pareto_real}
\end{figure}

Figure~\ref{fig:pareto_real} shows the validity in real settings, in which DCM-based optimization is comparable to the robot-based optimization ($N=20$) in terms of $I_{H}$.
The baseline method was not evaluated herein because it failed to predict the task objective $\ltask$ (Fig.~\ref{fig:pred_err_real}).
In Fig~\ref{fig:pareto_real}, DCM found compliant solutions (i.e., $\lcomp < 80$), but not the task objective-oriented solutions (i.e., $\ltask < 0.025$) unlike robot-based optimization.
This suggests taht DCM's prediction accuracy in real robot settings must be improved, which is the scope of future studies.

Table \ref{tab:summary_opt_cost} summarizes the optimization cost of the real robot experiment, suggesting that robot-free optimization performed by DCM was more efficient than robot-based (approximately three-fold faster).
In the experiment, a single trial was completed in approximately 4 min%
\footnote{
  The trial comprised input verification, rebuilding of a MATLAB control program, robot operation, and environment reset.
  The control program was rebuilt to apply a new stiffness and was the bottleneck of each trial.
}%
, thus robot-based optimization ($N=20$) was completed in 80 min.
In contrast, DCM conducted 1,000 trials virtually in robot-free optimization and found five Pareto solutions in only 5 min.
The accuracy of the obtained solutions was validated by conducting five robot tests, which was completed in 20 min.
Note that this validation can be eliminated if DCM can achieve sufficient prediction accuracy, leading to further optimization efficiency.

\begin{table}
  \centering
  \caption{Optimization cost in real robot experiments.}
  \label{tab:summary_opt_cost}
  \begin{tabular}{ccc|c} \toprule
    & Optimization & Validation & Total Time \\\midrule
    \multicolumn{3}{l|}{\textbf{robot-based optimization}~\cite{okada2023learning}} & \multirow{3}{*}{$\simeq 80$ min.} \\%\hline
    \# robot trials $N$ & 20 & - &  \\
    wall-clock time & $\simeq 80$ min. & - & \\
    \midrule
    \rowcolor[gray]{0.9}%
    \multicolumn{3}{l|}{\textbf{robot-free optimization by DCM (ours)} } & \\%\hline
    \rowcolor[gray]{0.9}%
    \# robot trials $N$ & 1,000 (virtual) & 5 & $\simeq 25$ min. \\
    \rowcolor[gray]{0.9}%
    wall-clock time & $\simeq 5$ min. & $\simeq 20$ min. & \\
    \bottomrule
  \end{tabular}
\end{table}

\section{Conclusion} \label{sec:conclusion}
This paper has proposed DCM for model-based learning of variable impedance control.
DCM predicts contact forces from given stiffness and demonstrated trajectories, realizing Bayesian optimization without trials on real robots.
To simulate highly nonlinear contact dynamics, diffusion models involving iterative calculations were used.
Using these models, inductive bias required for the contact simulation was incorporated.
The experiments on prediction accuracy evaluation and stiffness learning demonstrated the effectiveness of DCM in improving the accuracy and efficiency of stiffness learning.

Moreover, the similarity between DCM and the existing simulators in terms of optimization were discussed.
However, DCM is a completely black-box model, and its explainability was not analyzed herein.
Explainability analysis and/or the gray-boxing of DCM that incorporates more inductive biases from the analytical contact models will be performed in future research.

Improving the prediction accuracy of DCM particularly in real environments will also be targeted in future research.
Recently, substantial volumes of robot demonstrations have been collected to develop \textit{foundational models} for robot control~\cite{brohan2022rt,brohan2023rt}; however, these efforts were mainly focused on pick-and-place operations.
If a large dataset of contact-rich tasks can be built, it can be used to enhance the prediction accuracy of DCM.
This will allow us to use DCM as a foundational model for force control.

\appendix

\subsection{Ablation Study} \label{sec:ablation_study}
\begin{figure}[t]
  \centering
  \includegraphics[width=0.4\textwidth]{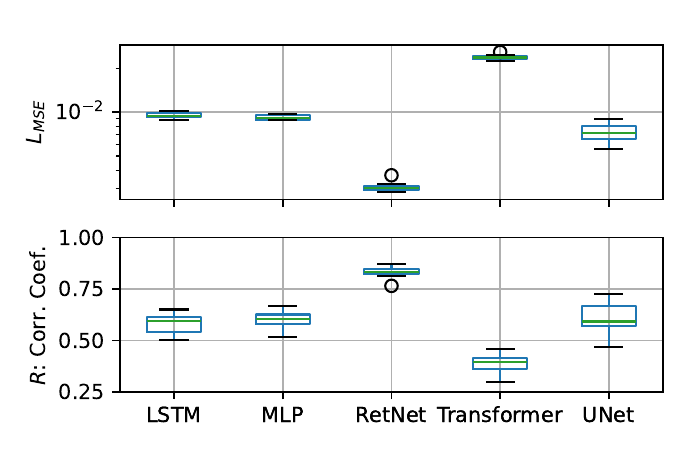}
  \vspace*{-5mm}
  \caption{
  Ablation study of model architecture of the score function $\epsilon_{\theta}(\cdot)$,
  in which RetNet exhibited the best performance.
  In addition to RetNet used in the main experiment, we evaluated the performance of other models such as long-short-time-memory (LSTM), U-Net, Transformer, and multi layer perceptron (MLP); MLP processes each timestep independently.
  }
  \label{fig:ablation_model}
\end{figure}

\begin{figure}[t]
  \centering
  \includegraphics[width=0.4\textwidth]{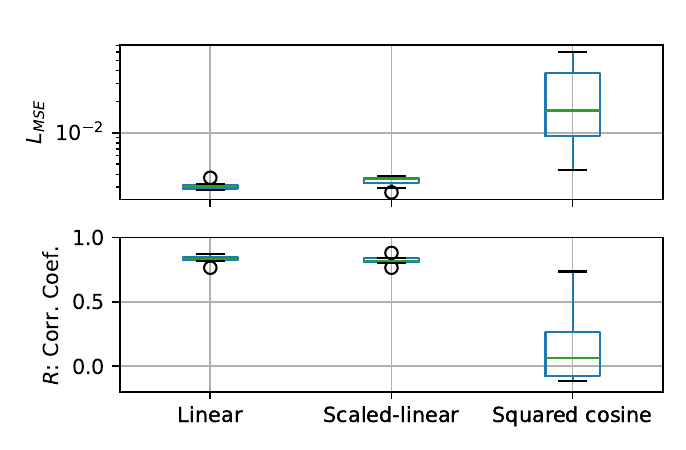}
  \vspace*{-5mm}
  \caption{
    Ablation study of noise schedules (i.e., variances $\beta_{1:T}$).
    Previous literature on robotics, wherein diffusion models were used~\cite{janner2022planning,chi2023diffusion} have reported that the squared cosine schedule~\cite{nichol2021improved} showed the best performance, but it did not work well in the force prediction task herein.
  }
  \label{fig:ablation_schedule}
\end{figure}

Figure~\ref{fig:ablation_model} and \ref{fig:ablation_schedule} show the results of the ablation study for the design decision of the score function $\epsilon_{\theta}(\cdot)$.

% \clearpage

\bibliography{iros2024}
\bibliographystyle{ieeetr}

\end{document}